\documentclass[a4paper,12pt]{article}
\usepackage{harvard}
\usepackage{url}
\usepackage{times}
\usepackage{endnotes}
\begin{document}

\title{\LARGE \bf Chasing the Ghosts of Ibsen: A computational stylistic analysis of drama in translation}
\author{ \large Gerard Lynch \& Carl Vogel \\
  \normalsize Computational Linguistics Group\\
  \normalsize  Department of Computer Science and Statistics\\
  \normalsize Trinity College\\
  \normalsize Dublin 2,Ireland\\
  \texttt{\{gplynch,vogel\}@tcd.ie}}

\date{June 22, 2009}

\maketitle

\section{Introduction}
Research into the stylistic properties of translations is an issue which has received some attention in computational stylistics. Previous work by \citeasnoun{Rybicki2006} on the distinguishing of character idiolects in the work of Polish author Henryk Sienkiewicz and two corresponding English translations using Burrow's Delta method concluded that idiolectal differences could be observed in the source texts and this variation was preserved to a large degree in both translations. This study also found that the two translations were also highly distinguishable from one another.

\citeasnoun{Burrows2002} examined English translations of Juvenal also using the Delta method, results of this work suggest that some translators are more adept at concealing their own style when translating the works of another author whereas other authors tend to imprint their own style to a greater extent on the work they translate.

Our work examines the writing of a single author, Norwegian playwright Henrik Ibsen, and these writings translated into both German and English from Norwegian, in an attempt to investigate the preservation of characterization, defined here as the distinctiveness of textual contributions of characters. 

\section{Background} 
Many studies in computational stylistics have focused on tasks which are related to those of authorship attribution but are not concerned with the notion of attributing authorship to texts of unknown provenance. 
A related area of study is the idea of \textit{pastiche}, an intended imitation of an author's style in the same language, which contrasts with translation as an intended imitation of an authors style but in a different language.
\citeasnoun{Somers2003} conducted experiments involving pastiche, the author in question was Lewis Carroll and the pastiche was a modern children's fable written by Gilbert Adair called \textit{Alice through the Needle's Eye} in which the author attempted to imitate the style of Carroll in such works as \textit{Through the Looking Glass} and \textit{Alice's Adventures in Wonderland}. Various techniques used in authorship attribution were used in the task, including methods of lexical richness, principal component analysis, the cusum technique\endnote{See \citeasnoun{Farr96} for a detailed explanation of the origins of this technique, including detailed examples of the method's use in a legal setting.}, and others. Some methods distinguished the pastiche from the original and some did not. \citeasnoun{Somers2003} conclude as follows: If a pastiche is indistinguishable from the original by an authorship attribution method, can it be said that the pastiche is in fact a perfect imitation of the original, or is it the case flawed?
In the case of translation which is of relevance to our current work, the question can be formulated in a different way: If a translation is highly similar stylistically to other works by the same translator, is the translation a faithful one?

This current study builds on previous work detecting character voices in the poetry of Irish poet Brendan Kennelly by \citeasnoun{VB07} and a study on characterization in playwrights by \citeasnoun{VogelLynch08}. These studies were concerned with the language used by authors in the creation of character. The tools used in this study were used in these previous studies.

\section{Experimental Setup}
For these experiments, three works by Henrik Ibsen were used, \textit{A Doll's House} (1879) \textit{Ghosts} (1881), and  \textit{The Master Builder} (1892)\endnote{For the English versions of the plays, the print versions are collected in \citeasnoun{Ibsen1890}, Sharp's translations can be found in \citeasnoun{Sharp1911}, the collected works of Ibsen in German are to be found in \citeasnoun{Ibsen1898} and the Norwegian collected works are found in \citeasnoun{Ibsen1957}} . The electronic versions of these plays were obtained from \textit{Ibsen.net}\endnote{ http://www.ibsen.net, last verified \today,  contains comprehensive information about Ibsen's life and work together with links to his plays in the original form and in translation.} and Project Gutenberg. The contributions of each character are extracted using PlayParser\endnote{A Java based tool designed for this purpose, \citeasnoun{VogelLynch07}, describes the creation and benchmarking of this particular program.}. All stage instructions are discarded in this step, leaving only the remaining character dialogue. The method decomposes all texts associated with a category (here, persona or play) into chunks of equal size. Pairwise similarity metrics are computed for all chunks. The metric is just the average chi-square computation of the difference in distribution between pairs
of files for each token appearing in either file. Different sorts of tokenization capture different linguistic features for which one might consider distributions within and across text categories.  If the pairwise similarity scores are rank ordered, then one can exploit the intuitions that a homogeneous category will have a smaller rank-sum than a heterogeneous one, and that arbitrary samples from a homogeneous category should be more like the rest of that category than alternative categories. The method also provides a way to measure degree of homogeneity, the number of samples who are more like the rest of the category can be measured against a baseline creating by random sampling. See \citeasnoun{VogelLynch08} for a more detailed account of the method.
\section{Experiments}
\subsection{First Experiment}
The first experiment seeks to compare character homogeneity over different languages. The second experiment compares two different translations of the same play in order to quantify similarity between parallel translations.
{\small
\begin{table}[htb]
\begin{center}
\begin{tabular}{|c|c|c|} \hline
Play&Language&Translator\\ \hline \hline
Gespenster(\textit{Ghosts)}&German&Sigurd Ibsen \\ \hline
Ein Puppenhaus(\textit{A Doll's House})&German&Marie Von Borch \\ \hline
Baumeister Solness(\textit{The Master Builder})&German&Marie Von Borch \\ \hline
The Master Builder&English&William Archer \& Edmund Gosse\\ \hline
A Doll's House&English&William Archer \\ \hline
Ghosts&English&William Archer\\ \hline
Ghosts&English&R Farquarson Sharp\\ \hline
\end{tabular}
\end{center}
\caption{Plays and Translators} \label{playtrans}
\end{table}
}
Table \ref{playtrans} shows the plays and their respective translators. As mentioned, the first 10k of text per character was examined and this was split into 5 sections. Thus, the criteria for inclusion in the study was that the character should contain at least 10k of text and 11 characters were examined, as detailed in Table \ref{charplays}. Only the version of \textit{Ghosts} translated by Archer is used in the first experiment. The results named in the next section have statistical significance.

{\small
\begin{table}[htb]
\begin{center}
\begin{tabular}{|c|c|} \hline
Character&Play\\ \hline \hline
Engstrand&Ghosts\\ \hline
Pastor Manders&Ghosts \\ \hline
Oswald&Ghosts \\ \hline
Mrs Alving&Ghosts \\ \hline
Helmer&A Dolls House \\ \hline
Krogstad&A Dolls House \\ \hline
MrsLinde&A Dolls House \\ \hline
Nora&A Dolls House \\ \hline
Aline&The Master Builder \\ \hline
Hilde&The Master Builder \\ \hline
Solness&The Master Builder \\ \hline
\end{tabular}
\end{center}
\caption{Characters and their plays} \label{charplays}
\end{table}
}

The results for the first experiment showed that character homogeneity varies to some extent over the translations, the character idiolects are not necessarily preserved to the same degree as the originals. When letter frequencies are measured, the Norwegian original language characters prove to be more homogeneous than the translations, examples include the character of Engstrand who is homogeneous in English and Norwegian but not German, however, one character whose language remains distinct across all of the translations is Nora, the heroine from \textit{A Doll's House} and one of the typical strong female characters found in Ibsen's drama.\endnote{Hedda Gabler being the other one who springs to mind, further studies may incorporate a wider range of plays and characters. }However, when the play is taken as the category, we find that the chunks of personas from each play are more similar to 
the personas from the same play than from different plays, and this is consistent across languages. So while within character homogeneity is not always preserved, the homogeneity of the plays remains relatively consistent across languages.
\section{The Second Experiment}
The second experiment sought to examine whether two translations of the same original text into the same language are distinguishable by translator as in the work by Rybicki which delineated the work by each, while observing the preservation of idiolect in each. The experimental setup was similar to the first experiment with the character contributions separated and split into five files each. This time, however, the characters from the two translations of \textit{Ghosts} by William Archer and Robert Farquharson Sharp were compared with each other.

Our findings were that the characters from Archer's translation were more homogeneous in general than those of Sharp's translation. Of the characters which were not homogeneous, the text segments were more similar to the segments of the same character by the corresponding author than any other writings by the same author. Sharp's characters tended to be more similar to the corresponding Archer character more often than vice versa. This suggests that both authors have managed to perform faithful translations which are not highly influenced by their own writing style. It also suggests that Sharp may have used Ibsen's translation as a reference when crafting his own.\endnote{It is not fully clear from any forewords to the e-texts when exactly the translations themselves were first published, however it does state that the first performance in English was in 1890, using Archers translation, Sharp's translations were first published in 1911, according to \newline \url{http://www.leicestersecularsociety.org.uk/library_shelf.htm}, last verified \today}

This result contrasts with \citeasnoun{Rybicki2006} who found that the two translations of Sienkiewicz separated cleanly from one another with a preservation of individual character idiolects. However, Rybicki makes clear that the two English translations were done almost one hundred years apart with the second translator taking specific steps to bring the language of Sienkiewicz into the 20th century. Also, we are aware
that results between the studies of two different authors are not directly comparable and do not seek to draw definite parallels, merely to reflect on related work in the same sphere.
\section{Conclusion}
In this research, character idiolects in translation have been examined. Future work will involve using different metrics for comparison along with comparing different selections of text from the characters considered, along with the comparisons of translations of different authors by the same translator.
\theendnotes
\bibliographystyle{apsr} 
\bibliography{DH2009}

\end{document}